# Handwriting recognition and automatic scoring for descriptive answers in Japanese language tests

Hung Tuan Nguyen[1][0000-0003-4751-1302], Cuong Tuan Nguyen[1][0000-0003-4751-1302], Haruki Oka[2*], Tsunenori Ishioka[3][0000-0003-0267-5653], and Masaki Nakagawa[1][0000-0001-7872-156X]

[1] Tokyo University of Agriculture and Technology, Tokyo, Japan
[2] Recruit Co. Ltd., Tokyo, Japan
[3] The National Center for University Entrance Examinations, Tokyo, Japan
`fx7297@go.tuat.ac.jp, fx4102@go.tuat.ac.jp,`
`haruki_oka@r.recruit.co.jp, tunenori@rd.dnc.ac.jp,`
`nakagawa@cc.tuat.ac.jp`

**Abstract.** This paper presents an experiment of automatically scoring handwritten descriptive answers in the trial tests for the new Japanese university entrance examination, which were made for about 120,000 examinees in 2017 and 2018. There are about 400,000 answers with more than 20 million characters. Although all answers have been scored by human examiners, handwritten characters are not labeled. We present our attempt to adapt deep neural network-based handwriting recognizers trained on a labeled handwriting dataset into this unlabeled answer set. Our proposed method combines different training strategies, ensembles multiple recognizers, and uses a language model built from a large general corpus to avoid overfitting into specific data. In our experiment, the proposed method records character accuracy of over 97% using about 2,000 verified labeled answers that account for less than 0.5% of the dataset. Then, the recognized answers are fed into a pre-trained automatic scoring system based on the BERT model without correcting misrecognized characters and providing rubric annotations. The automatic scoring system achieves from 0.84 to 0.98 of Quadratic Weighted Kappa (QWK). As QWK is over 0.8, it represents an acceptable similarity of scoring between the automatic scoring system and the human examiners. These results are promising for further research on end-to-end automatic scoring of descriptive answers.

## 1 Introduction

Descriptive answers are better for evaluating learners' understanding and problem-solving ability. They encourage learners to think rather than select. However, scoring them requires large work and time. In recent years, it was proposed to add descriptive

---

questions in the new university entrance common examinations in Japan as well as the current multiple-choice questions [1], but given up due to the short period of scoring handwritten answers and the anxiety about reliable scoring.

One approach is to score handwritten descriptive answers automatically and feedback scores to examinees and examiners to correct scoring errors. Another approach is to apply automatic scoring or to cluster them for human examiners to score them efficiently and reliably [2, 3]. For both of them, handwritten answers need to be recognized and scored.

A few datasets storing handwritten answers have been published and used in research on handwriting recognition, such as SCUT-EPT (Chinese handwritten answers) [4] and Dset-Mix, which is artificially prepared by synthesized handwritten math answers [5]. Note that these datasets are all fully labeled and ideally suited to train handwriting recognizers based on deep neural networks.

The National Center for University Entrance Examinations (NCUEE) conducted trial tests for the new university entrance common exams with 64,518 and 67,745 examinees in 2017 and 2018, respectively. Three descriptive questions were included in the Japanese language test for each trial test. All handwritten answers were scored by human examiners. The scanned images and scores by human are used in this research.

However, the offline images are only raw images and have not been segmented or labeled. It is infeasible to prepare labels for a whole set of scanned handwritten answers. On the other hand, automatic pattern recognition methods, especially well-known deep neural networks, require large-scale labeled data for training. Hence, we present normalization, segmentation, and handwriting recognition from this handwritten answer set and their automatic scoring. In particular, we focus mainly on training the handwriting recognizer to adapt to the actual data from the examinee's answers. In addition, we also incorporated the language model to re-rank the predicted candidates so that the ambiguous patterns are corrected by linguistic context. In summary, the main contributions are as follows:

- We present an ensemble deep neural network-based recognizer for offline Japanese handwritten answer recognition.
- We propose a training procedure with multiple steps (pre-training, fine-tuning and ensemble learning) to adapt the ensembled handwriting recognizer to real patterns.
- We evaluate the handwriting recognizer in combination with the latest automatic scoring system.

Note that the combined architecture does not try to correct of misrecognized characters or require rubric annotations since we expect that some small misrecognitions would not affect to automatic scoring model. In practice, the human examiners do not need to recognize all the characters in answers as they score answers based on the meaning of sentences or detection of keywords.

## 2 Related Works

Handwritten text recognition has been studied for many decades [6]. Offline recognition (Optical recognition from scanner or camera recently) followed by online recognition from tablet have been developed for reading postal addresses, bank checks, business forms and documents. Recently, more challenging problem of historical document recognition is being studied extensively [7, 8].

However, there is only a few studies on handwritten text recognition for handwritten answers. Handwritten answers are usually different from each other in terms of content on top of writing style so that the cost for completely labeling them is almost impractical. Recently, one of the first large-scale datasets of examination answers, the SCUT-EPT dataset, was introduced [4]. It contains 50,000 handwriting Chinese text line images provided by 2,986 volunteers.

The proposed SCUT-EPT dataset shows challenges, including character erasure, text line supplement, character/phrase switching, noised background, non-uniform word size, and unbalanced text length. The current advanced text recognition methods, such as convolutional recurrent neural network (CRNN), exhibits poor performance on the proposed SCUT-EPT dataset. According to the visualizations and error analyses, the authors reported that the performance of examiners is much better than the CRNN method. Furthermore, they compared three sequential labelling methods, including connectionist temporal classification (CTC), attention mechanism, and cascaded attention-CTC. Even though the attention mechanism has been demonstrated to be effective in English scene text recognition, its performance was far inferior to the CTC method in the case of the SCUT-EPT dataset with a large-scale character set. This paper presents the first attempt to recognize and score large-scale unconstrained Japanese handwritten answers from high school students for the trial university entrance examinations.

## 3 Handwritten Japanese Answer Dataset

In 2017 and 2018, the National Center for University Entrance Examinations of Japan has conducted trial tests using descriptive questions for the university entrance examinations for the Japanese language and Math. At that time, nearly 38% of high schools in Japan attended these trial tests, but the students were not asked to participate in all subjects. For the Japanese language tests, 64,518 and 67,745 answer sheets were collected in 2017 and 2018, respectively. We call this dataset NCUEE-Handwritten Japanese Answers (NCUEE-HJA) with two sub-datasets NCUEE-HJA-2017 and NCUEE-HJA-2018.

Note that all answer images are not labeled except their scores, which is a major barrier to adopting the latest deep neural network-based handwriting recognition methods. Thus, we need to prepare a small but effective number of samples for training handwriting recognizers. In order to prepare these labeled samples, we first employed multiple segmentation stages to obtain the single characters from the answer images, as shown in **Fig. 1**. Then, the text-line label is manually prepared.

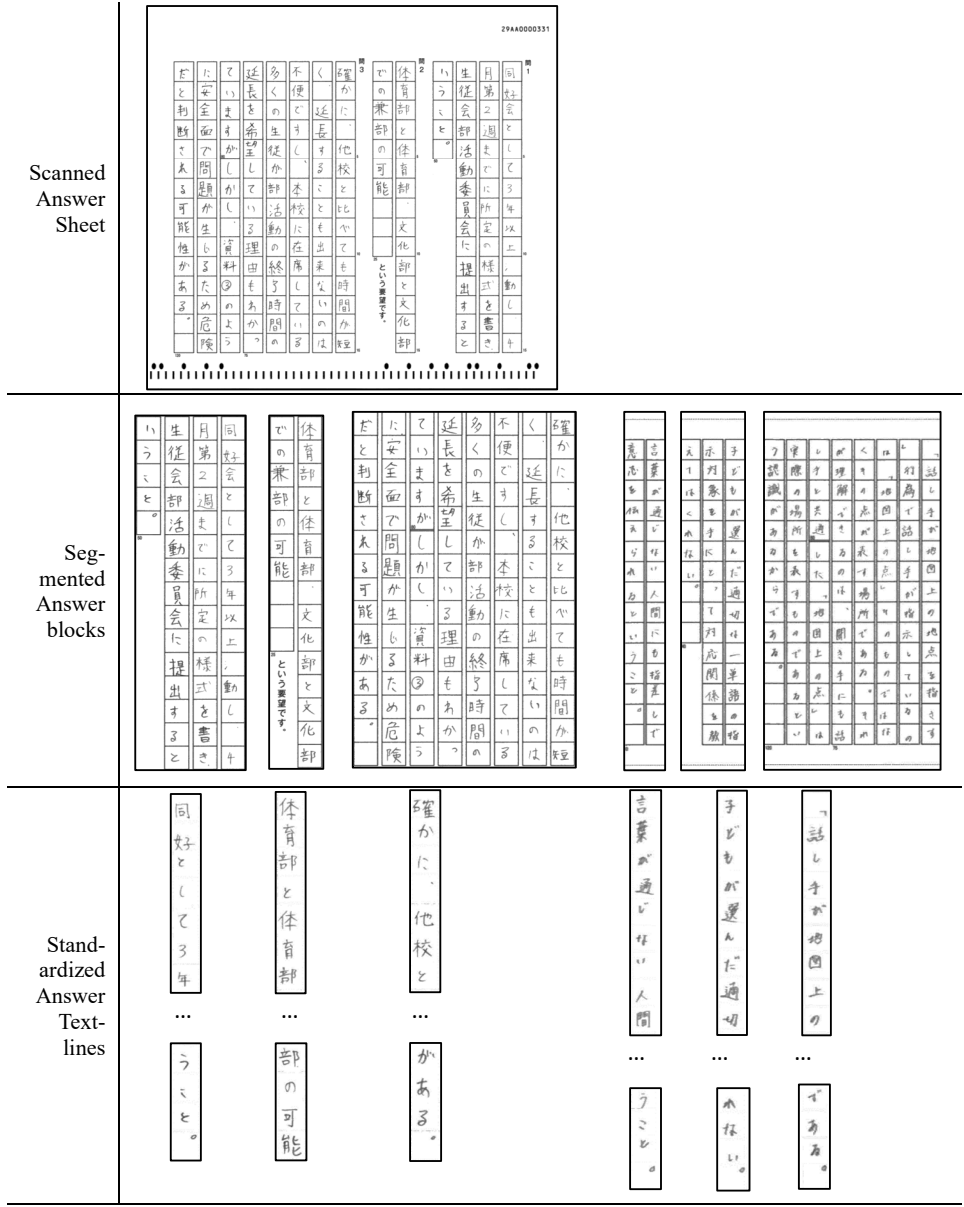

Scanned Answer Sheet

Segmented Answer blocks

Standardized Answer Text-lines

(a) NCUEE-HJA-2017      (b) NCUEE-HJA-2018

**Fig. 1.** Samples in the NCUEE-HJA dataset and their processed results.

### 3.1 Handwritten Text-line Segmentation

We used a pipeline of two steps to segment handwritten text-lines from answer sheet images. First, the answer block segmentation step is completed by horizontal and vertical histogram-based projections to detect answer blocks. For the NCUEE-HJA-2017 dataset, an answer image of an examinee consists of three answers for three descriptive questions, as shown in the top row of **Fig. 1**, and it is segmented into answer blocks as shown in the second row of **Fig. 1**. Note that the answer sheet of the NCUEE-HJA-2018 dataset has been pre-segmented so that the above answer block segmentation stage is not necessary.

Secondly, individual characters are segmented. We apply morphological operators to detect horizontal and vertical borderlines to form character boxes. Then, they are removed. Next, each character image inside each box is padded to a unique size of 64-by-64 using white pixels. Finally, a sequence of white-pixel padded character images are concatenated in the order from top right to bottom left for generating a vertically handwritten text-line, as shown in the third row of **Fig. 1**. The accuracy of the pipeline for handwritten text-line segmentation is 99.42% on the NCUEE-HJA dataset, where the unsuccessful segmented images have heavy noises caused by ground color stain made with eraser.

### 3.2 Splitting and Labeling samples

We manually labeled 100 handwritten answers for each dataset (NCUEE-HJA100), equivalent to 0.05% of the NCUEE-HJA dataset. We performed a semi-automatically labelling process for other 1,000 handwritten answers for each dataset (NCUEE-HJA1K). In the semi-automatically labelling process, a small CNN model was trained on NCUEE-HJA100 and then deployed on NCUEE-HJA1K.

Next, the predicted labels on NCUEE-HJA1K were manually verified. Although this process was costly, it was much faster than manually assigning labels to prepare an adequate number of patterns for training deep neural networks.

### 3.3 Statistics

As shown in **Table 1**, the NCUEE-HJA dataset is larger than the SCUT-EPT dataset in terms of the number of writers and the number of samples. Its number of categories is less than one of SCUT-EPT, but it consists of many character types such as alphabet, kanji, hiragana, katakana, and numeric. Moreover, the SCUT-EPT dataset has a large number of labeled patterns while the NCUEE-HJA dataset only has a large number of unlabeled patterns, which raises the most challenging problem in pattern recognition. Thus, the proposed method is different from the previous studies, where many labeled patterns are available.

**Table 1.** Comparison between SCUT-EPT and NCUEE-HJA.

| Database \ Characteristics | SCUT-EPT | NCUEE-HJA | |
|---|---|---|---|
| Language | Chinese | Japanese | |
| Collected Year | 2018 | 2017 | 2018 |
| Type | Offline Handwritten Text (labeled) | Offline Handwritten Text (unlabeled) | |
| No. of Categories | 4,250 | 3,125 | |
| No. of Writers | 2,986 (2,986) | 64,518 (1,000) | 67,745 (1,000) |
| No. of Text-lines | 50,000 (40,000) | 193,554 (1,000) | >11 million (61,883) |
| No. of Characters | 1,267,167 (1,018,432) | 203,235 (1,000) | >10 million (53,370) |

( ) *presents the value of the corresponding training set.*

## 4 Handwritten Answer Recognition and Automatic Scoring

Deep Neural Networks (DNNs) have achieved state-of-the-art results on pattern recognition challenges such as object detection, image classification, or even image transcription during the last decade. In this paper, we propose employing the different DNN models for two tasks: handwriting recognition and text classification, to perform automatic handwritten answer scoring as shown in **Fig. 2**.

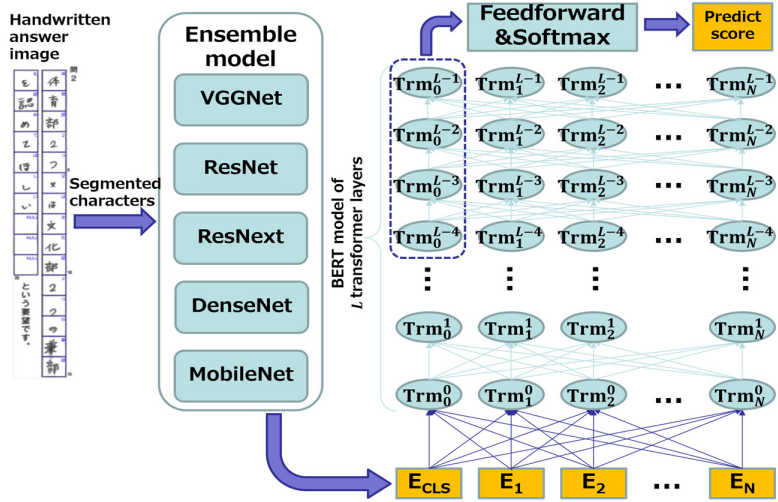

**Fig. 2.** Overall of the pipeline for Handwritten Answer Recognition and Automatic Scoring.

### 4.1 Handwritten Answer Recognition

Generally, deep neural networks have become a state-of-the-art approach for image classification, especially Convolutional Neural Networks (CNNs). According to the results from [7], we employ a set of five well-known CNN models (VGG16, MobileNet24, ResNet34, ResNeXt50, and DenseNet121). These models are modified to adapt to the small input images as a single character image of 64-by-64 pixels. Their first convolutional layers are modified to use the kernel size of 3-by-3. Moreover, the final classification layers are modified to match the number of categories in the NCUEE-HJA dataset. The details of CNN models are presented as follows.

The first network is derived from VGGNet, which consists of 16 convolutional layers divided into four blocks, as presented in [9]. The second network, MobileNet24, with 24 convolutional layers [10], is proposed to work efficiently on smart devices with limited computational resources. The third and fourth networks based on the residual connections (ResNet34 and ResNeXt50) consist of 34 and 50 layers. These layers are grouped into multiple blocks with 3 convolutional layers and a residual connection in each block [11, 12]. The fifth network is DenseNet121, in which the extracted feature maps in lower-level layers are concatenated to obtain dense features for higher-level layers [13]. We reused the pre-trained parameters of the five models on ImageNet as provided in the PyTorch vision library† to have a generalized initialization.

For solving the challenge of a few labeled samples, good initialization of recognition models is important. It requires a large-scale dataset in a similar domain to the NCUEE-HJA dataset. Thus, we employed all handwritten character datasets from the ETL database [14]. We applied various transformations to training patterns from the ETL database to generalize the training data, such as rotation, shearing, shifting, blurring, and noising with random parameters.

The pre-trained models using the ETL database are finetuned on the verified labeled NCUEE-HJA1K. Note that the different trained models are prepared for each year due to the difference in image quality of each year (NCUEE-HJA-2018 has lower quality than NCUEE-HJA-2017 using the JPEG lossy compression). Next, these finetuned models are evaluated on the NCUEE-HJA test sets with 100 manually labeled answers. An ensemble model of all five models is also evaluated by averaging the results from the single models.

For post-processing, we employed a simple language model, n-gram, with the beam-search decoder in order to reorder the predicted candidates according to linguistic context. In the experiments, we compute a 5-gram model from a large text corpus of Japanese Wikipedia using the KenLM method [15].

### 4.2 Automatic Scoring

For automatic scoring, the predicted text is classified into some ranks, such as [0, 1, 2, 3], which correspond to the scores assigned by examiners. Thus, automatic scoring is considered as text classification in this paper. We use BERT [16], which is pre-trained on Japanese Wikipedia. By adding the information of the handwritten answers to it, we can make reasonable estimates using the knowledge obtained from the large-scale model. The procedure is as follows:

1. Divide the recognized answers into morphemes. Morphemes are called tokens, and each is given an ID. A special token called CLS is given at the beginning of the sentence.
2. Convert the ID assigned to each token into a 768-dimensional vector from the result of pre-learning in Japanese Wikipedia.
3. Obtain the vector of CLS tokens in the final 4 layers of the 12 hidden layers in BERT.
4. Combine the extracted 768-dimensional vectors in the column direction. This creates a 768×4 matrix.
5. Dimension transformation of this matrix with a linear layer is then applied to the softmax function to output the category with the highest probability.

---

† https://pytorch.org/vision/stable/models.html

For the estimation by BERT, all the data were divided into 3:1:1, that is, 60%, 20%, 20%, used as training data, validation data, and test data, respectively. Adam is used for model optimization [17], and training was performed with a batch size of 16 and an epoch number of 5. Finally, the second-order weighted $\kappa$ coefficient (QWK) [18] is used to measure the degree of agreement between scores by the automatic scoring system and those by human examiners. Generally, QWK values of 0.61–0.8 indicate substantial agreement, and >0.8 suggest almost perfect agreement [19].

## 5 Experiment Results

The following sections present the performance of the recognition model and the automatic scoring model.

### 5.1 Performance of Recognition Model

The performance of the shallowest and deepest models as well as the ensemble of all five models, are shown in **Table 2**. Although the image quality of NCUEE-HJA in 2018 is lower than in 2017, the character recognition rates are improved due to fine-tuning and the n-gram context procession. Thus, the CNN models were able to extract discriminative features regardless of the image quality after the fine-tuning process. Moreover, even the shallowest model performed similarly to the deepest one. However, a single model might not perform well on a specific dataset due to the distribution of features and the network structure. In order to avoid this problem, we employed the ensemble by averaging the predictions from single models. Note that the ensemble model does not outperform the VGG16 and DenseNet121 on the test sets of NCUEE-HJA 2017 and 2018, respectively. However, it is expected to perform well in a large number of writing styles.

Table 2. Character Recognition Accuracy (%) on the NCUEE-HJA-2017 and NCUEE-HJA-2018 test sets of 100 answers.

|  | Fine-tuned | Applied n-gram | NCUEE-HJA-2017 | NCUEE-HJA-2018 |
|---|---|---|---|---|
| **VGG16** | ✗ | ✗ | 67.36 | 39.50 |
|  | ✗ | ✓ | 74.55 | 49.06 |
|  | ✓ | ✗ | 98.11 | 98.61 |
|  | ✓ | ✓ | **98.52** | 98.27 |
| **Dense Net121** | ✗ | ✗ | 67.23 | 49.21 |
|  | ✗ | ✓ | 73.87 | 53.71 |
|  | ✓ | ✗ | 94.00 | 98.02 |
|  | ✓ | ✓ | 96.08 | **98.44** |
| **Ensemble of five models** | ✗ | ✗ | 74.64 | 54.97 |
|  | ✗ | ✓ | 84.15 | 65.92 |
|  | ✓ | ✗ | 96.99 | 98.18 |
|  | ✓ | ✓ | <u>97.75</u> | <u>97.97</u> |

## 5.2 Performance of Automatic Scoring Model

**Table 3** shows the statistics for scoring a total of 6 questions, 3 for each of 2017 and 2018. The number of answers, the score range, the average, the standard deviation, the number of allowed characters, and the QWK scores of different models are shown in order. We calculated the mean and standard deviation of the scores from a 4-point scale {a, b, c, d} by setting a→3, b→2, c→1, d→0.

From the score range and mean columns in Table 3, 2017#1 seems most straight-forward, as implied by its highest mean score, while 2017#3 seems most difficult due to the lowest mean score. According to QWK scores of these questions, the worst case of 2017#3 is 0.82; while the worst cases of other questions are higher than 0.9. Thus, QWK scores seem to be correlated with the difficulty of the question. As mentioned in the recognition performance analysis, the ensemble model is expected to perform well on a large number of writing styles. In **Table 3**, the ensemble model outperforms the single models, such as VGG16 and DenseNet121, on all questions.

Table 3. Statistics on scoring for each question.

| Questions | No. of answers | Score range | Mean (Std. Div.) | # of allowed char. | Quadratic Weighted Kappa (QWK) | | | |
|---|---|---|---|---|---|---|---|---|
|  |  |  |  |  | VGG 16 | Dense Net 121 | Ensemble model (no n-gram) | Ensemble model |
| **2017 #1** | 62,222 | 0~6 | 4.46 (1.67) | ~50 | 0.977 | 0.974 | 0.975 | **0.980** |
| **2017 #2** | 61,777 | 0~2 | 1.51 (0.87) | ~25 | 0.957 | 0.952 | 0.957 | **0.959** |
| **2017 #3** | 59,791 | 0~5 | 0.43 (1.10) | 80~120 | 0.844 | 0.820 | **0.847** | 0.830 |
| **2018 #1** | 67,332 | 0~3 | 2.51 (0.88) | ~30 | 0.972 | 0.970 | **0.973** | 0.970 |
| **2018 #2** | 66,246 | 0~3 | 1.87 (1.14) | ~40 | 0.952 | **0.953** | 0.950 | **0.953** |
| **2018 #3** | 58,159 | 0~3 | 0.76 (1.07) | 80~120 | 0.933 | 0.935 | 0.937 | **0.941** |

In order to investigate the robustness of our method with misrecognized text, we compare the ensemble model that combines fine-tuned models and uses n-gram with two fine-tuned single recognizers with n-gram. Moreover, the fine-tuned ensemble models with and without n-gram are compared. **Table 3** also shows the degree of deterioration. Although the recognition results are not perfect as 100%, the predicted text seems usable for automatic scoring. For the single models, the QWK scores decreased by nearly 3 percentage points (pcps). For the ensemble model without a language model, the QWK scores are almost similar to the ensemble model with an n-gram language model except for the question 2018#3. Even when considering the language model, the performance did not improve. We found that the more difficult the question, the larger gap in the QWK scores between the worst and best cases.

## 6 Conclusions

This paper presented the first attempt to score handwriting descriptive answers for Japanese language tests automatically. The proposed pipeline consists of multiple state-of-the-art deep neural networks to recognize Japanese handwriting characters and automatically score the answers. The recognizers achieved a high character accuracy of over 97%, with only 0.1% labeled patterns from the dataset. Moreover, the automatic scoring model performed almost the same as examiners with QWK scores from 0.84 to 0.98, depending on the difficulty of the questions. These results suggest that an end-to-end automatic scoring system on descriptive answers, where the recognizer and automatic scoring models are combined as a single model, would be promising for further research.

## Acknowledgement

This research is being partially supported by JSPS KAKENHI: Grant Number JP20H04300 and A-STEP: JST Grant Number JPMJTM20ML.